\documentclass[11pt,a4paper]{article}
\usepackage[hyperref]{acl2020}
\usepackage{times}
\usepackage{latexsym}

\usepackage{booktabs} 
\usepackage{longtable,tabu}
\usepackage{booktabs,arydshln} 

\usepackage{textcomp}

\usepackage{graphicx}

\usepackage{tikz}
\usetikzlibrary{shapes.geometric, arrows}
\usepackage{algorithm,algorithmic}
\usepackage{amsmath}
\usepackage{amssymb}
\makeatletter

\def\adl@drawiv#1#2#3{%
        \hskip.5\tabcolsep
        \xleaders#3{#2.5\@tempdimb #1{1}#2.5\@tempdimb}%
                #2\z@ plus1fil minus1fil\relax
        \hskip.5\tabcolsep}
\newcommand{\cdashlinelr}[1]{%
  \noalign{\vskip\aboverulesep
           \global\let\@dashdrawstore\adl@draw
           \global\let\adl@draw\adl@drawiv}
  \cdashline{#1}
  \noalign{\global\let\adl@draw\@dashdrawstore
           \vskip\belowrulesep}}
\makeatother

\usepackage{microtype}

\aclfinalcopy 

\title{Reduce Indonesian Vocabularies with an Indonesian Sub-word Separator}

\author{Mukhlis Amien \and Chong Feng \and Heyan Huang \\
        Beijing Institute of Technology \\ Beijing, China \\
\texttt{fengchong@bit.edu.cn} \\}

\date{}

\begin{document}
\maketitle
\begin{abstract}
Indonesian is an agglutinative language since it has a compounding process of word-formation. Therefore, the translation model of this language requires a mechanism that is even lower than the word level, referred to as the sub-word level. This compounding process leads to a rare word problem since the number of vocabulary explodes. We propose a strategy to address the unique word problem of the neural machine translation (NMT) system, which uses Indonesian as a pair language. Our approach uses a rule-based method to transform a word into its roots and accompanied affixes to retain its meaning and context. Using a rule-based algorithm has more advantages: it does not require corpus data but only applies the standard Indonesian rules. Our experiments confirm that this method is practical. It reduces the number of vocabulary significantly up to 57\%, and on the English to Indonesian translation, this strategy provides an improvement of up to 5 BLEU points over a similar NMT system that does not use this technique.

\end{abstract}

\section{Introduction}


In some language, for example, Indonesian, Turkish and German language has agglutination and compounding process of word formation, some language suffers a different degree of this processes than the other. So the translation model of this language requires a mechanism that is even lower than the word level, referred to as the sub-word level. In Indonesian, one of these rare word problems is caused by the compounding process of the  words due to the high word formation processes. It causes the same root word considered as another word by the vocabulary list. \par 

Since the invention of transformers \cite{vaswani2017attention}, Neural Machine Translation (NMT) has become state of the art and the latest approach in machine translation techniques \cite{DBLP:journals/corr/BahdanauCB14}\cite{DBLP:journals/corr/JeanCMB14}\cite{DBLP:journals/corr/SutskeverVL14}\cite{DBLP:journals/corr/ChoMGBSB14}\cite{kalchbrenner13emnlp}. NMT has shown promising results compared to traditional approaches such as Statistical Machine Translation (SMT)\cite{koehn2003}. A significant disadvantage in conventional NMT systems is their reduced ability to translate uncommon words, on the other hand, standard phrase-based systems such as statistical machine translation (SMT) system \cite{koehn2007}\cite{chiang2007}, do not suffer rare word problems at the same level because they can support more extensive vocabulary, and because they use explicit alignments, and phrase tables that allow the system to remember translations of infrequent words. \par

\begin{table}[ht!]

\centering
\resizebox{\columnwidth}{!}{%
\begin{tabular}{@{}llll@{}}
\hline
\textbf{Root-Word} & \textbf{compounding form} & \textbf{ Separation Form} & \textbf{Translation} \\ 
\hline

Makan(eat) & Makan-makan & prl$\sim$ makan & eating out \\
 & memakan & me$\sim$ makan & eat up \\
 & memakani & me$\sim$ makan $\sim$i & feed \\
 & Memakankan & me$\sim$ makan $\sim$kan & give something to eat \\
 & Makanan & makan $\sim$an & food \\
 & Dimakan & di$\sim$ makan & eaten \\
 & Pemakan & pe$\sim$ makan & eater \\
 & Termakan & ter$\sim$ makan & eaten accidentally \\
 & Sepemakan & se$\sim$ pe$\sim$ makan & the same time as people eat \\
 & Makan-makanan & prl$\sim$ makan $\sim$an & various kinds of food \\
     \hline

 Jalan(street) & berjalan-Jalan & ber$\sim$ prl$\sim$ jalan & having fun on foot \\
 & berjalan & ber$\sim$ jalan & walk \\
 & menjalani & me$\sim$ jalan $\sim$i & undergo \\
 & menjalankan & me$\sim$ jalan $\sim$ & do (duty, obligation, work) \\
 & jalanan & jalan $\sim$an & the street \\
 & pejalan & p$\sim$ jalan & pedestrian \\
 & perjalanan & pe$\sim$ jalan $\sim$an & to travel \\
 & sejalan & se$\sim$ jalan & in line \\

\end{tabular}%

}
\caption{Two examples of various form of "makan" and "jalan" root-word and separation Algorithm Result}
\label{tab:makandanjalan}
\end{table}

Motivated by the characteristic of Indonesian, which naturally causes this rare words problem, we propose a novel approach to address the rare word problem of the NMT system that using Indonesian as a pair language. Our approach is transforming word become its root and its accompanied affixes to retain its meaning and context. For instance, in table \ref{tab:makandanjalan}, the term "makan" and "jalan" in Indonesian should be a common word, but because it contains at least 10 word variations, every vocab of that variation is considered a separate vocab, therefore the phrase "sepemakan" is called OOV (Out Of Vocabulary) when it is not. Although this problem can be solved by machine learning such as the BPE technique \cite{BPE2020}, sentencepiece \cite{sentencepiece}, or unigram language model \cite{kudo2018subwordUnigram}, the Indonesian case will be easier to solve using a rule base, because the compounding process of word formation has clear and relatively unambiguous rules unlike English.



Our experiments result is to confirm that this approach is practical. On the English, to Indonesian translation task, this approach provides an improvement of up to 5 (with 32000 of most used vocabularies) BLEU points over a similar NMT system that does not use this technique.



\subsection{Problem Formulation}
The characteristics of Indonesian are many agglutination and compounding process in word formation; this lead to the translation using the word level approach will become ineffective due to the vocabulary that shouldn't be rare becomes rare. It creates a unique word problem, which is the main problem of NMT systems as previously investigated by Luong \cite{luong2015}. And then we propose a novel sub-word approach designed to overcome the characteristics of Indonesian.


\subsection{Contributions}
We have two contributions of this paper: first is datasets collection is done semi-manually and massively ( about 3.5 millions language pairs in Indonesian, English, and Mandarin). This collection attempts are still growing, that never been done before and can be downloaded\footnote{https://goo.gl/vdrW6u} in sqlite3 format. The second contribution is a novel Indonesian sub-word separation algorithm for data preparation in Indonesian pairs for NMT task with very notable improvement in BLEU score. The code accompanied this paper can be downloaded at Github\footnote{https://github.com/neimasilk/amien\_stemming}.

\section{Related Work}
\subsection{Indonesian is an Agglutinative Languages} \label{sub:indo}
Languages which is use agglutination widely are called agglutinative languages \cite{bodmer1972}. An example of such language is Indonesian as a Family of Austronesian Language, where for example, the Indonesian word "\textit{mempertanggungjawabkannya}", or meanings is "account for it", it consists of the morphemes \textit{mem-per-tanggung-jawab-kan-nya}. An Indonesian usually consists of words in which most of them not use the base dictionary entry. Words cause it does not stand alone but are often consist of compounding of prefixes, suffixes, infixes, and sometimes accompanied by possessive pronouns and particles \cite{derwin2014}. To overcome this problem in information retrieval, we usually use the word stemming technique. Stemming method is the process of finding the base word entry (root word) from a word form \cite{mirna2007}. Until this far, no effort has been made to develop a word separation of affixes from its root words for Indonesian. Instead, some works on stemming, which has gained more attention in its development for Indonesian. From all the published journals related to this topic, Indonesian researcher only develope stemming methods for the Indonesian language. Stemming aims to reduce the numbers of variation from a language to a standard, canonical representation (known as the stem). Indonesian stemming methods use root word as its stem; which means that mostly they are dictionary dependent. The stemming process may be different, according to the nature of the language itself. Indonesian is a morphologically complex language where almost every word can be turned with affixes. \par
According to Mirna \cite{mirna2007} the general rules of affix in Indonesian are:
\begin{equation}
\textbf{[[[DP+]DP+]DP+] root-word [[+DS][+PP][+P]]}
\label{rule}
\end{equation}
DP : Derivational Prefixes; DS : Derivational Suffixes; PP : Possessive pronouns; P: Particles. \\
\begin{table}[]
\centering
\resizebox{\columnwidth}{!}{%
\begin{tabular}{@{}ll@{}}
\toprule
Prefixes & Suffix \\ \midrule
"me$\sim$", "per$\sim$", "ber$\sim$", "ter$\sim$", and "di$\sim$" & "$\sim$kan" \\
"me$\sim$", "per$\sim$", "ter$\sim$", and "di$\sim$" & "$\sim$i" \\
"ber$\sim$" and "ke$\sim$" & "$\sim$an" \\ \bottomrule
\end{tabular}%
}
\caption{The common prefixes and suffixes combinations in Indonesian. The
prefixes "se-" and "pe-" are not in the list.}
\label{tab:combination1}
\end{table}
Prefixes in Indonesian produce derivative word of the root word \cite{asian2007}. These prefixes are complicated because some prefixes can vary depending on the first letter of the root word, and the first letter of the root word may also be eliminated or modified depending on the prefix it is connected to the word. The prefixes are "\textit{pe-}","\textit{me-}", "\textit{se-}", "\textit{ter-}", "\textit{di-}", "\textit{ber-}", and "\textit{ke-}" . Additional two prefixes:  "\textit{kau-}" and "\textit{ku-}" are also recognized as prefixes although they are less formal and not commonly used, so in this proposed algorithm we did not include prefixes "\textit{kau-}" and "\textit{ku-}". \par
The suffix do not change the shape of the root word. According to the \textit{Tata Bahasa Baku Bahasa Indonesia (TBBBI)}  "A Standard Grammar of Indonesian" by Moeliono and Dardjowidjojo \cite{TBBBI}, there are just three suffixes in Indonesian, specifically "\textit{-an}","\textit{-i}",and "\textit{-kan}" .  In Indonesian, there are particles and possessive suffixes connected at the end of a root word that is not counted as suffixes grammatically, but in algorithm they are the same as suffixes so the rule of suffixes can be applied. These particles and possessive suffixes are connected to the word but do not alter the root words. However, they change the meaning of the root word. There are three possessive suffixes in Indonesian, i.e "\textit{-ku}", "\textit{-mu}", and "\textit{-nya}", meaning possession by first, second, and third-person respectively. Examples are "\textit{pensilku}" or my pencil, "\textit{pensilmu}" is your pencil, and "\textit{pensilnya}" is his/her pencil. The suffix "\textit{-nya}"  are also can be applied for the possessive of the third person plural. As reporting by the TBBBI \cite{TBBBI}, the particles "\textit{-lah}", "\textit{-kah}", and "\textit{-tah}" do not alter the root words. For example, the words "\textit{makanlah}" (please eat) and "\textit{diakah?}" (is it you?), which stem from "\textit{makan}" (eat) and "\textit{dia}" (him/her), do not alter after being attached to the particles. The particle "\textit{-tah}" is now out-of-date and never used in modern Indonesian language. \par
It is possible to produce a new word by combining more than one prefix, more than one suffix, and an infix into a root word or a repeated word. Joining these combination affixes still adheres to the rules of adding their element affixes. As table \ref{tab:combination1} shows the combinations of prefixes and suffixes that occur frequently. Table \ref{table_combination_2} is a list of some prefixes-suffixes pairs that never resemble together.
The Mirna stemming algorithm \cite{mirna2007} is based on comprehensive morphological rules that group together and encapsulate allowed affixes combination (table \ref{tab:combination1}) and disallowed affixes combination (table \ref{table_combination_2}), including prefixes, suffixes, and confixes (the combination of prefixes and suffixes), which are also known as circumfixes. Affixes can be inflectional or derivational \cite{payne_describing_1997}. This classification of affixes leads to the rules \ref{rule}.
\begin{table}[]

\centering
\begin{tabular}{@{}ll@{}}
\toprule
\textbf{Prefix} & \textbf{Disallowed suffixes} \\ \midrule
"ber$\sim$" & "$\sim$i" \\
"di$\sim$" & "$\sim$an" \\
"ke$\sim$" & "$\sim$i" and "$\sim$kan" \\
"me$\sim$" & "$\sim$an" \\
"ter$\sim$" & "$\sim$an" \\
$\sim$per$\sim$" & "-an" \\ \bottomrule
\end{tabular}
\caption{Some of prefixes and suffixes combinations in Indonesian that are never appear together.}
\label{table_combination_2}

\end{table}

\subsection{Tokenization and Sub-word Tokenization} \label{sub:token}
Tokenization is crucial for text processing tasks such as sentiment analysis, topic identification, and spam filtering. In text categorization, sentence representation can be calculated based on sentence construction tokens, i.e., a sentence is converted into smaller meaningful units such as letters, words, and sub-words. The token can then be modeled using a neural network, such as that used in Neural Machine Translation (NMT).
In general, there are two tokenization methods: the machine learning method and the rule-based method. For the machine learning technique, numerous tokenization models, including sentencepiece \cite{sentencepiece}, BPE \cite{BPE2020}, and the unigram language model \cite{kudo2018subwordUnigram}, are frequently employed. As for the rule-based approach, it is typically custom-tailored, whereas the tokenization approach is dependent on the language. The most common and simplest approach is the whitespace separation approach between tokens, which is so simple because all that is required is to separate tokens based on space characters. Because Indonesian has a consistent and unambiguous compound form, rule-based sub-word tokenization will deliver good results without requiring a large training set. We only require a sufficient dataset for evaluation.

\subsection{Rare Word Problems in NMT System}

The rare word problem on machine translation is still an open problem \cite{DBLP:journals/corr/KoehnK17}. A vocabulary of neural models is usually limited to about 20000 to 50000 words. However, a machine translator is an open vocabulary problem. Some language, for example, Indonesian has agglutination and compounding process of word formation. So the translation model of this language requires a mechanism that is even lower than the word level. As an example, consider such as the Indonesian word of ("\textit{keberuntunganmulah}" $\rightarrow$"\textit{ke-ber-untung-an-mu-lah}") a segmented variable-length representation is more make sense than encoding the word as a fixed-length vector. For Word level NMT models, the translation of OOV words has been addressed to a dictionaries look-up \cite{DBLP:journals/corr/JeanCMB14}\cite{DBLP:journals/corr/LuongPM15}, this is reasonable assumptions, but that often does not hold in practice. For example, there is not always a 1 to 1 correspondence between source-target words due to variance within the degree of morphological synthesis between language. Also, word-level models cannot translate or generate unseen words. Sennrich\cite{subword2016} has proposed the sub-word model using bit pair encoding (BPE) technique, and his analysis shows that the neural networks can learn compounding and transliteration from sub-word representations. However, the Sennrich method is not entirely suitable for Indonesian because of the changing form of root word caused by the presence of affixes.

\subsection{Neural Machine Translation (NMT)}
In this work of experimentation, we use the model of Luong et al. (2015) \cite{DBLP:journals/corr/LuongPM15}, which uses a D-LSTM to encode the input sequence and a separate D-LSTM to create the translation. Then the encoder reads the source sentence, one word at a time also produces a huge vector that represents the whole source sentence. The decoder is initialized by this vector and generates a translation, one word at a time until it emits the end-of-sentence symbol $<$eos$>$.

The technique we propose applies to almost all NLP tasks using Neural networks as a basis, such as sentiment analysis,  question and answering, summarization, and machine translation. However, in this case, we use neural machine translators as benchmarks. 

\subsection{Evaluation: BLEU and Perplexity}
To determine if there has been an improvement in the translation of the NMT, we employ two measures, BLEU \cite{BLEU} and Perplexity \cite{perplexity}, which are both commonly used evaluation tools.\par
BLEU, or the Bilingual Evaluation Understudy, is a score that compares a potential text translation to one or more reference translations. Although designed for translation, it can also be used to analyze text output for a variety of applications involving natural language processing. \citet{BLEU} proposed the BLEU score.  The method works by comparing n-grams in the candidate translation to n-grams in the source text, where a 1-gram or unigram comparison would be each token and a bigram comparison would be each word pair. The comparison is conducted irrespective of the word order.\par
In general, perplexity \cite{perplexity} measures the accuracy with which a probability model predicts a sample. Language models can be evaluated using perplexity in the context of NLP. A language model is a probability distribution over sentences: it is capable of both generating believable human-written sentences (if it is a competent language model) and evaluating the quality of already written sentences. A decent language model should be able to assign a higher probability to a well-written text than to a poorly written document; it should not be "confused" when faced with a well-written document. Thus, the perplexity measure in NLP is a technique to quantify a model's 'uncertainty' in predicting text.

\section{Proposed Approach}
\subsection{Detail Methodology}



Rare word problems are well known issues for languages that have complex morphology (e.g. agglutinative) such as Indonesian. This rare word problem can be caused by a real rare word, such as a unique person's name, but there is also a rare word problem caused by an agglutinating type of language. Our approach is to solve the rare word problem caused by the nature of Indonesian which is a type of agglutinating language. As in the  chapter \ref{sub:token}, there are two approaches commonly used by researchers, namely the machine learning approach or the rule based approach. Our approach uses a rule-based method to transform a word into its roots and accompanied affixes to retain its meaning and context. Using a rule-based algorithm has more advantages: it does not require corpus data but only applies the standard Indonesian rules.

As mentioned in chapter \ref{sub:indo}, Indonesian has many words variation, as in the example in table \ref{tab:makandanjalan}, the word "\textit{makan}" (eat) and "\textit{jalan}" (street), can become many vocabulary variations. Therefore, in this paper, we will develop a method to separate an Indonesian word into its root word and affix without changing the original meaning; this is to reduce the number of vocabulary significantly. This algorithm is derive from a modified Mirna stemming technique \cite{mirna2007}. The stemming process is beneficial for information retrieval, but this process can change the meaning of the sentences from its context. The idea is we can separate word and become a combination of affixes (suffixes, prefixes, and infixes) and its root-word, and encode it with tilde ($\sim$) symbol, so when we combine it again, it can be combined and become the original word. For examples: \textit{sepemakan $\rightarrow$ se$\sim$ pe$\sim$ makan} and after combination process: \textit{se$\sim$ pe$\sim$ makan $\rightarrow$ sepemakan}, this process looks simple, but due to changing of the root-word form, this algorithm becomes more complicated because the separate algorithm should match with the rule of Affixes stripping in table \ref{tab:separation_rule_ber_ter}, table \ref{tab:separation_rule_me}, and table \ref{tab:separation_rule_pe}. \par
Separating the Indonesian word affixes is relatively challenging. There are variations of affixes, including prefixes, infixes, suffixes, and confixes. Furthermore, Indonesian has repeated words, combinations of affixes, and combinations of affixes with repeated words. Indonesian also has compound words that are written together when attached to a prefix and a suffix.


\subsection{Proposed Algorithm}
\begin{figure}[htbp]
\centerline{\includegraphics[width=0.5\columnwidth]{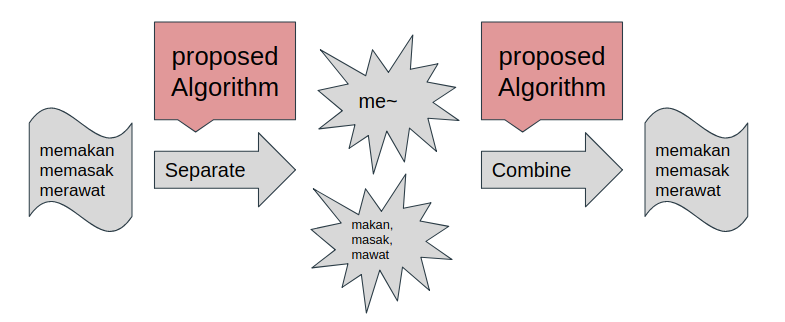}}
\caption{Process of separation method}
\label{fig:1}
\end{figure}
Figure \ref{fig:1} is a simple overview of separation and combination algorithm. Function separate() accepts input words and detect if there are any affixes, and then split into its subwords. After being put into the NMT system then all subwords are combined into words by identifying all tilde ($\sim$) symbols. The algorithm \ref{alg:1} begins by reading a sentence and then dividing it into word. Then each word is compared whether it has root-word or not. The process continued with the next word until the last sentence. If the word has a root-word, then a further process will be carried out, namely the word separation process as in algorithm \ref{alg:2}. The algorithm \ref{alg:2} consists of the process of separation of prefixes, root-word, and suffixes, and then possessive pronouns attached to the end of the word. Then the word is decoded into sub-word and put it into the main sentence that has been tokenized (example below). Example: \\
Input = \textit{Benarkah semua korban gempa Aceh sudah terjamin kebutuhan pokoknya?} \\
Output = \textit{benar $\sim$kah semua korban gempa aceh sudah ter$\sim$ jamin ke$\sim$ butuh $\sim$an pokok $\sim$nya ?} \par
\begin{table}[]
\resizebox{\columnwidth}{!}{%
\begin{tabular}{@{}llll@{}}
\toprule
\multicolumn{1}{l}{\textbf{No.}} & \multicolumn{1}{l}{\textbf{Affix}} & \multicolumn{1}{l}{\textbf{Separation Rule}} & \multicolumn{1}{l}{\textbf{Example}}\\ \midrule

1 & ber$\sim$ & $ber\sim V... | be\sim rV...$ & berencana $\rightarrow$ ber$\sim$ rencana \\
\cdashlinelr{1-4}
2 & ber$\sim$ & $ber\sim CAP...\ where\ C!='r'\ and\ P!='er'$ & berhasil $\rightarrow$ ber$\sim$ hasil \\
\cdashlinelr{1-4}
3 & ber$\sim$ & $ber\sim CAerV...\ where\ C!='r'$ & bebercak $\rightarrow$ ber$\sim$ bercak \\
\cdashlinelr{1-4}
4 & ber$\sim$ & $bel\sim ajar...$ & belajar $\rightarrow$ ber$\sim$ ajar \\
\cdashlinelr{1-4}
5 & ber$\sim$ & $be\sim C1erC2...\ where\ C1!=\{'r'\ |\ 'l'\}$ & beterbangan $\rightarrow$ ber$\sim$ terbang $\sim$an \\ \midrule

6 & ter$\sim$ & $te\sim rV...\ $& terendah $\rightarrow$ ter$\sim$ rendah \\
\cdashlinelr{1-4}
7 & ter$\sim$ & $ter\sim CerV...\ where\ C!='r'$ & terjerumus $\rightarrow$ ter$\sim$ jerumus \\
\cdashlinelr{1-4}
8 & ter$\sim$ & $ter\sim CP...$ & tersisa $\rightarrow$ ter$\sim$ sisa \\ \bottomrule

\end{tabular}%
}
\caption{Prefix separation rule set for ber$\sim$, ter$\sim$ }
\label{tab:separation_rule_ber_ter}

\end{table}

\begin{table}[]

\resizebox{\columnwidth}{!}{%
\begin{tabular}{@{}llll@{}}
\toprule
\multicolumn{1}{l}{\textbf{No.}} & \multicolumn{1}{l}{\textbf{Affix}} & \multicolumn{1}{l}{\textbf{Separation Rule}} & \multicolumn{1}{l}{\textbf{Example}}\\ \midrule

1 & me$\sim$ & $me\sim \{l|r|w|y\}V...$ &
\begin{tabular}[c]{@{}l@{}}  
melebihi $\rightarrow$ me$\sim$ lebih $\sim$i \\
meraih $\rightarrow$ me$\sim$ raih \\
mewujudkan $\rightarrow$ me$\sim$ wujud $\sim$kan \\
meyakini $\rightarrow$ me$\sim$ yakin $\sim$i
\end{tabular} \\
\cdashlinelr{1-4}
2 & me$\sim$ & $mem\sim \{b|f|v\}...$ &
\begin{tabular}[c]{@{}l@{}}  
membedakan $\rightarrow$ me$\sim$ beda $\sim$kan \\
memfasilitasi $\rightarrow$ me$\sim$ fasilitas $\sim$i \\
memviralkan $\rightarrow$ me$\sim$ viral $\sim$kan 
\end{tabular} \\
\cdashlinelr{1-4}
3 & me$\sim$ & $mem\sim pe... $& mempertahankan $\rightarrow$ me$\sim$ pe$\sim$ tahan $\sim$kan  \\
\cdashlinelr{1-4}
4 & me$\sim$ & $me\sim m\{rV|V\}...\ |\ me\sim p\{rV|V\}...$ &
\begin{tabular}[c]{@{}l@{}}  
memukul $\rightarrow$ me$\sim$ pukul \\
memprakarsai $\rightarrow$ me$\sim$ prakarsa $\sim$i \\
memerkosa $\rightarrow$ me$\sim$ perkosa 
\end{tabular} \\
\cdashlinelr{1-4}
5 & me$\sim$ & $men\sim \{c|d|j|z\}...$ &
\begin{tabular}[c]{@{}l@{}}  
mencoba $\rightarrow$ me$\sim$ coba \\
mendapat $\rightarrow$ me$\sim$ dapat \\
menjadi $\rightarrow$ me$\sim$ jadi \\ 
menzalimi $\rightarrow$ me$\sim$ zalim $\sim$i
\end{tabular} \\
\cdashlinelr{1-4}
6 & me$\sim$ & $me\sim nV...\ |\ me\sim tV...$ &
\begin{tabular}[c]{@{}l@{}}  
menilai $\rightarrow$ me$\sim$ nilai \\
menulis $\rightarrow$ me$\sim$ tulis 
\end{tabular} \\
\cdashlinelr{1-4}
7 & me$\sim$ & $meng\sim \{g|h|q|k\}...$ &
\begin{tabular}[c]{@{}l@{}}  
menggunakan $\rightarrow$ me$\sim$ guna $\sim$kan  \\
mengharapkan $\rightarrow$ me$\sim$ harap $\sim$kan  \\
mengqisash $\rightarrow$ me$\sim$ qisash \\ 
mengkalkulasi $\rightarrow$ me$\sim$ kalkulasi
\end{tabular} \\
\cdashlinelr{1-4}
8 & me$\sim$ & $meng\sim V...\ |\ meng\sim kV...$ &
\begin{tabular}[c]{@{}l@{}}  
menganggap $\rightarrow$ me$\sim$ anggap  \\
mengasihi $\rightarrow$ me$\sim$ kasih $\sim$i
\end{tabular} \\
\cdashlinelr{1-4}
9 & me$\sim$ & $meny\sim sV...$ & menyelamatkan $\rightarrow$ me$\sim$ selamat $\sim$kan \\
\cdashlinelr{1-4}
10 & me$\sim$ & $mem\sim pV...\ where\ V!='e' $& memikirkan $\rightarrow$ me$\sim$ pikir $\sim$kan \\ \bottomrule
\end{tabular}
}
\caption{Prefix separation rule set for me$\sim$ }
\label{tab:separation_rule_me}
\end{table}
\begin{table}[]

\resizebox{\columnwidth}{!}{%
\begin{tabular}{@{}llll@{}}
\toprule
\multicolumn{1}{l}{\textbf{No.}} & \multicolumn{1}{l}{\textbf{Affix}} & \multicolumn{1}{l}{\textbf{Separation Rule}} & \multicolumn{1}{l}{\textbf{Example}}\\ \midrule

1 & pe$\sim$ & $pe\sim \{w\}V... $& pewakaf $\rightarrow$ pe$\sim$ wakaf  \\
\cdashlinelr{1-4}
2 & pe$\sim$ & $per\sim V...\ |\ pe\sim rV...$ &
\begin{tabular}[c]{@{}l@{}}  
perairan $\rightarrow$ pe$\sim$ air $\sim$an   \\
peraih $\rightarrow$ pe$\sim$ raih
\end{tabular} \\
\cdashlinelr{1-4}
3 & pe$\sim$ & $per\sim CAP...\ where\ C!='r'\ and\ P!='er'$  & perbuatannya $\rightarrow$ pe$\sim$ buat $\sim$an $\sim$nya  \\
\cdashlinelr{1-4}
4 & pe$\sim$ & $pem\sim \{b|f\}...$ &
\begin{tabular}[c]{@{}l@{}}  
pembunuhan $\rightarrow$ pe$\sim$ bunuh $\sim$an   \\
pemfaktoran $\rightarrow$ pe$\sim$ faktor $\sim$an
\end{tabular} \\
\cdashlinelr{1-4}
5 & pe$\sim$ & $pen\sim \{c|d\}...$ &
\begin{tabular}[c]{@{}l@{}}  
pencapaian $\rightarrow$ pe$\sim$ capai $\sim$an   \\
pendidik $\rightarrow$ pe$\sim$ didik
\end{tabular} \\
\cdashlinelr{1-4}
6 & pe$\sim$ & $pe\sim nV...\ |\ pe\sim tV...$  &
\begin{tabular}[c]{@{}l@{}}  
penasehat $\rightarrow$ pe$\sim$ nasehat  \\
penabur $\rightarrow$ pe$\sim$ tabur
\end{tabular} \\
\cdashlinelr{1-4}
7 & pe$\sim$ & $peng\sim \{g|h|k\}...$ &
\begin{tabular}[c]{@{}l@{}}  
penggelapan $\rightarrow$ pe$\sim$ gelap $\sim$an   \\
penghargaan $\rightarrow$ pe$\sim$ harga $\sim$an \\
pengkultusan $\rightarrow$ pe$\sim$ kultus $\sim$an  
\end{tabular} \\
\cdashlinelr{1-4}
8 & pe$\sim$ & $peng\sim V...$ & pengakuan $\rightarrow$ pe$\sim$ aku $\sim$an   \\
\cdashlinelr{1-4}
9 & pe$\sim$ & $penye\sim sV...$ & penyesalan $\rightarrow$ pe$\sim$ sesal $\sim$an   \\
\cdashlinelr{1-4}
10 & pe$\sim$ & $pe\sim lV...\ except:\ 'pelajar'\ return\ 'pe\sim \ ajar' $&
\begin{tabular}[c]{@{}l@{}}  
pelumas $\rightarrow$ pe$\sim$ lumas   \\
pelajar $\rightarrow$ pe$\sim$ ajar 
\end{tabular} \\ \bottomrule

\end{tabular}
}
\caption{Prefix separation rule set for pe$\sim$ }
\label{tab:separation_rule_pe}
\end{table}
In table \ref{tab:separation_rule_ber_ter}, table \ref{tab:separation_rule_me} and table \ref{tab:separation_rule_pe} are the rule of how Indonesian prefix change its root-word shape, V stands for a vowel (a, i, u, e, o), C stands for consonant, A represents any alphabet character (a-z), and P represents a short fragment of words, such as 'er'.
\tikzstyle{startstop} = [rectangle, rounded corners, minimum width=3cm, minimum height=1cm,text centered, draw=black, fill=red!30]
\tikzstyle{io} = [trapezium, trapezium left angle=70, trapezium right angle=110, minimum width=3cm, minimum height=1cm, text centered, draw=black, fill=blue!20, text width=7em, text badly centered, node distance=2cm, inner sep=0pt]
\tikzstyle{process} = [rectangle, minimum width=3cm, minimum height=1cm, text centered, draw=black, fill=blue!20, node distance=2cm, inner sep=0pt]
\tikzstyle{decision} = [diamond, draw, fill=blue!20, 
   text width=4.5em, text badly centered, node distance=2.5cm, inner sep=0pt]
\tikzstyle{arrow} = [thick,->,>=stealth]
\tikzstyle{bulat}= [circle,fill=black,inner sep=0pt,minimum size=3pt]

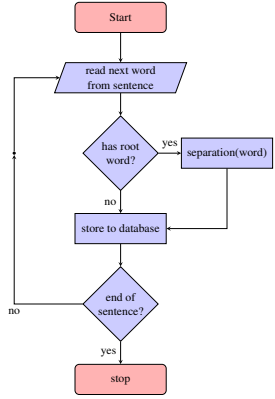
\begin{figure}[htbp]
\centering
\scalebox{0.4}{
\begin{tikzpicture}[node distance=2cm, auto]
\node (start) [startstop] {Start};
\node (in1) [io, below of=start] {read next word from sentence};
\node (pilihan1) [decision, below of=in1] {has root word?};
\node (bulatan)[bulat, left of = pilihan1, node distance =3.5cm]{};
\node (separate) [process, right of=pilihan1, node distance=3.5cm]{separation(word)};
\node (store)[process, below of=pilihan1, node distance=2.5cm] {store to database};
\node (pilihan2) [decision, below of=store, node distance=2.5cm] {end of sentence?};
\node (selesai)[startstop, below of=pilihan2, node distance=2.5cm] {stop};

\draw [arrow] (start) -- (in1);
\draw [arrow] (in1) -- (pilihan1);
\draw [arrow] (pilihan1) -- node[anchor=south] {yes} (separate);
\draw [arrow] (pilihan1) -- node[anchor=east] {no} (store);
\draw [arrow] (separate) |- (store);
\draw [arrow] (store) -- (pilihan2);
\draw [arrow] (pilihan2) -| node[anchor=north] {no} (bulatan);
\draw [arrow] (bulatan) |- (in1);
\draw [arrow] (pilihan2) -- node[anchor=east] {yes} (selesai);

\end{tikzpicture}
}
\caption{Flowchart of the separation process}
\label{fig:flowchart_detail}
\end{figure}

\begin{center}
\scalebox{0.75}{
    \begin{minipage}{\linewidth}
\begin{algorithm}[H]

 \caption{Algorithm for sentence separation}
 \label{alg:1}
 \begin{algorithmic}[1]
 \renewcommand{\algorithmicrequire}{\textbf{Input:}}
 \renewcommand{\algorithmicensure}{\textbf{Output:}}
 \REQUIRE \textit{sentence} \\ \textbf{require:} function \textit{separate\_word()}
 \ENSURE  \textit{separated\_sentence} \# all words in the sentence that has root word, will be separated and lowered. \\ \textbf{Example input:} \textit{"Benarkah semua korban gempa Aceh sudah terjamin kebutuhan pokoknya?"}   , \\ \textbf{example output:} \textit{"benar \texttildelow kah semua korban gempa aceh sudah ter\texttildelow  \ jamin ke\texttildelow \ butuh \texttildelow an pokok \texttildelow nya?"}
 \\ \textit{Initialisation} :
  \STATE sentence $\leftarrow$ to\_lower(sentence)
  \STATE words[] $\leftarrow$ sentence.split(' ')
  \STATE stems $=$ []
 \\ \textit{LOOP Process:}
  \FOR {word in words}
  \STATE stem\_result = separate\_word(word)
  \STATE stems.append(stem\_result)
  \ENDFOR
  \STATE separated\_sentence = ' '.join(stems)
 \RETURN separated\_sentence
 \end{algorithmic}
 \end{algorithm}
 \end{minipage}
 }
 \end{center}

\begin{center}
\scalebox{0.75}{
    \begin{minipage}{\linewidth}
\begin{algorithm}[H]
 \caption{Algorithm for word separation}
 \label{alg:2}
 \begin{algorithmic}[1]
 \renewcommand{\algorithmicrequire}{\textbf{Input:}}
 \renewcommand{\algorithmicensure}{\textbf{Output:}}
 \REQUIRE \textit{word} \\ \textbf{require:} function \textit{stem\_word()}, function \textit{encode\_prefix()}, function \textit{encode\_suffix()}
 \ENSURE  \textit{separated\_word} \#separated word by it sub-word unit. \\ \textbf{Example input:} \textit{"sepemakan"}   ,\\ \textbf{example output:} \textit{"se\texttildelow \ pe\texttildelow \ makan"}
 \\ \textit{Initialisation} :
  \STATE stem\_word $\leftarrow$ stem\_word(word)
  \STATE prefix $\leftarrow$ encode\_prefix(word)
  \STATE suffix $\leftarrow$ encode\_suffix(word)
  \STATE separated\_word $\leftarrow$ prefix+'\texttildelow \ '+stem\_word+'\texttildelow \ '+suffix
 \RETURN separated\_word 
 \end{algorithmic}
 \end{algorithm}
 \end{minipage}
 }
 \end{center}



\section{Experiment}
\subsection{The Data Sets Collection}

Considering the quality of the translation of the NMT system is determined by the amount of data. Thus, the quality and quantity of training data are influential on the translation results. Our focus of this task is to collect as many data as possible, without reducing data quality. However, due to the difficulty of finding parallel corpus in Indonesian, English, and Mandarin, the data set will be a mixture of the manual interpretation done by human and then mixed with automatic translation with google translate. For example, a manual parallel corpus of Indonesia-English, combined with automatic conversion of Mandarin, and then English-Mandarin Manual parallel corpus, mixed with automatic translation of Indonesian. And some of the monolingual corpus for examples Wikipedia Indonesia, English and Mandarin, combined with automatic conversion of the two respected languages as shown in table \ref{tab:3}. \\
\begin{table}[]

\centering
\resizebox{\columnwidth}{!}{%
\begin{tabular}{@{}ll@{}}
\toprule
Data Source & Quantity \\ \midrule
\begin{tabular}[c]{@{}l@{}}Wikipedia Indonesia \\ Google  Translate (ID-EN-ZH)\end{tabular} & 1000000 \\ \midrule
\begin{tabular}[c]{@{}l@{}}Wikipedia Mandarin (Simplified) \\ Google Translate (ZH-ID-EN)\end{tabular} & 350000 \\ \midrule
CASICT (ZH EN)-Google Translate (ID-EN) & 2300000 \\ \midrule
NEU ( ZH-EN) - Google Translate (ID-EN) & 2000000 \\ \midrule
Kompas.com - Google Translate (ID-EN-ZH) & $\sim$ \\ \bottomrule
\end{tabular}
}
\caption{Data sets Source}
\label{tab:3}
\end{table}
Table \ref{tab:4} is a dataset statistics. The process of collecting this dataset is still work in progress. The latest dataset can be downloaded\footnote{https://goo.gl/vdrW6u} in sqlite3 format.
\begin{table}[]

\resizebox{\columnwidth}{!}{%
\begin{tabular}{@{}lll@{}}
\toprule
\textbf{Statistics} & \textbf{Quantity} & \textbf{Description} \\ \midrule
ID-EN-ZHCN & 3,426,608 Pairs & 
\begin{tabular}[c]{@{}l@{}}Parallel corpus in \\
 trilingual:Indonesia, \\  
 English and \\
 Mandarin\\
 \end{tabular} \\ \midrule
\begin{tabular}[c]{@{}l@{}}Average of word count per\\ sentencee\end{tabular} & \begin{tabular}[c]{@{}l@{}}22.33 Indonesia, \\ 25.58 English, \\ 27.78 Mandarin\end{tabular} & \begin{tabular}[c]{@{}l@{}}
We counting sum of \\
every word in  \\
sencences and divided \\ 
with total sentences
\end{tabular} \\ \midrule

\end{tabular}%
}
\caption{Datasets statistics}
\label{tab:4}
\end{table}

\subsection{Experiment Framework}
\begin{figure}[htbp]
\centerline{\includegraphics[width=0.7\columnwidth]{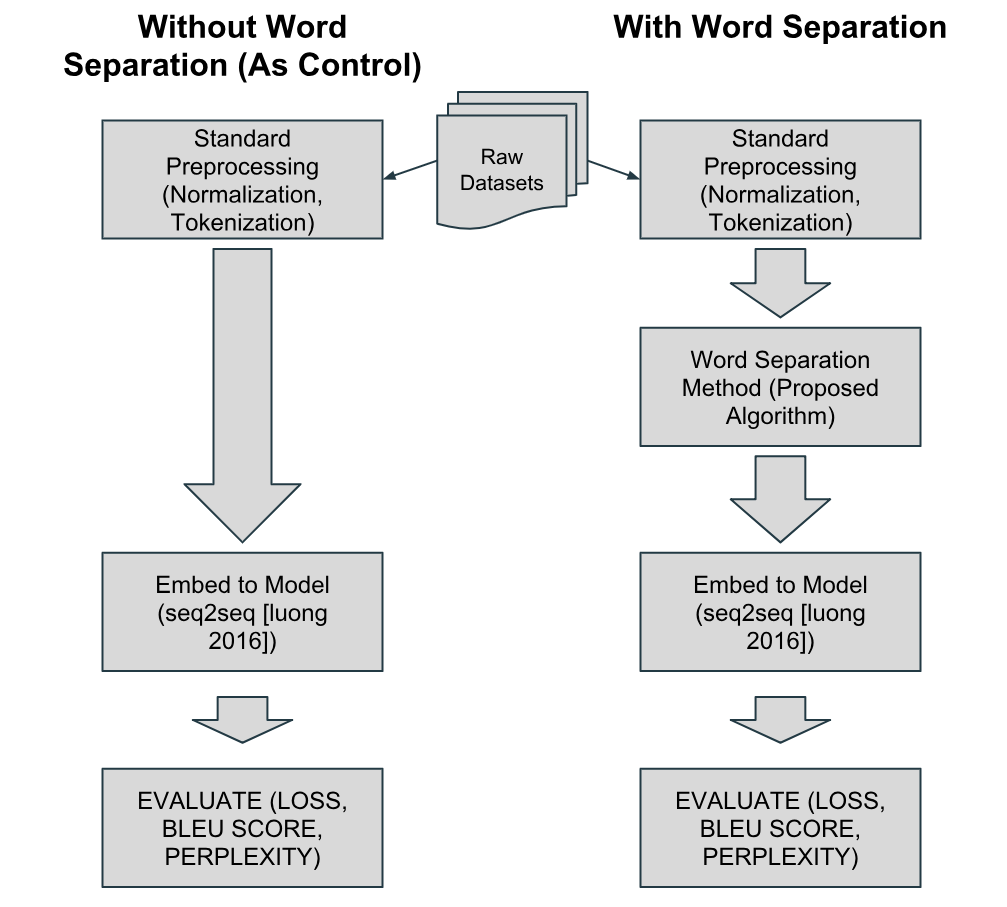}}
\caption{Experiment Framework in General}
\label{fig:3}
\end{figure}

Figure  \ref{fig:3} is an overview of the experiment. Begins with standard preprocessing in NLP techniques, such as lowering all capital letters, tokenization, cleaning unwanted symbols. Next, at the same data set, we conduct two experiments. First, is a baseline experiment without using any additional method. And the second experiment is using the algorithm we propose, the separation-combination method. The next process is training seq2seq technique proposed by Luong \cite{DBLP:journals/corr/LuongPM15}. When we wrote this paper, Luong's approach was state of the art. Then from the results of these two experiments, we will compare the translation quality. The translation outcome is evaluated using the BLEU score and perplexity level.





\section{Result and Analysis}
\subsection{Rare Words Reduction}


After separation, it will recalculate the number of vocabulary list in the dataset to discover the reduction of the words counts. To learn words decrease, we compare the results with the dataset before being separated. From table \ref{tab:5}, there is a significant decrease in the total number of a vocabulary list. There was a reduction of 57.26\% of the entire vocabulary count compared with before separation.


\begin{table}[]

\centering
\resizebox{\columnwidth}{!}{%
\begin{tabular}{@{}lll@{}}
\toprule
\textbf{No.} & \textbf{Description} & \textbf{Quantity} \\ \midrule
1. & \begin{tabular}[c]{@{}l@{}}Unique tokens of the vocabularies \\ in training data set before applying \\ separation algorithm\end{tabular} & 1,925,245 tokens \\ \midrule
2. & \begin{tabular}[c]{@{}l@{}}Unique tokens of the vocabularies\\  in training data set after applying \\ separation algorithm\end{tabular} & 822,875 tokens \\ \midrule
3. & Difference in Vocabulary reduction & 1,102,370 tokens \\ \midrule
4. & \textbf{Vocabularies Reduction percentage} & \textbf{57.26\%} \\ \bottomrule
\end{tabular}
}
\caption{Comparation of vocabularies counting before and after separation}
\label{tab:5}
\end{table}

\subsection{Comparison of Word Reduction using Our Approach with Existing Methods}


\subsection{NMT Result Improvement}
Here we implement the Luong \cite{DBLP:journals/corr/LuongPM15} model to compare the translation results between datasets without using the separation algorithm with using the algorithm. From figure \ref{fig:4}, the BLEU score has a significant increase, reaching 5 points, from 40 to 45. This improvement proves that the process of separation greatly influences the translation results.
From the log perplexity data \cite{perplexity} in figure \ref{fig:5}, it also decreasing from 2.6 to 0.65, this shows that the model is confident with the translation results.  In our case, the development perplexity reaches its peak on the last iteration then starts getting converge, by stopping training on the 1st million iterations, the log perplexity improved a slight bit to 0.65, but this didn't make a big difference in BLEU score improvement.

Table \ref{translate} in  Appendix A  are three examples of English-Indonesian translations using our method and the baseline method. "Our result" is the result of interpretation after the separation process, and the "baseline result" is the translation result before the separation process. The "origin (EN)" is English origin sentences and "truth ref (ID)" is from our testing data sets. BLEU score is an individual BLEU value from translation results.

\section{Conclusion}

We have two main contributions of this paper. First, data sets collection is done semi-manually and massively about 3.5 millions language pairs and growing in three pairs (Indonesian, English, Mandarin) That never been done before and can be downloaded freely. The second contribution is a novel Indonesian sub-word separation method for reducing rare Indonesian words. \par
The goal of the paper is to show that our algorithm approach enhances the NMT system. Our method is improving NMT towards reaching state-of-the-art, proving its promising results. Research on Indonesian vocabulary reduction caused by the agglutination process is a new approach. The nearest approach is the stemmer method. Reduction of vocabulary caused by sub-word has a significant effect on the translation results. With the technique in this paper, there was a significant reduction in the vocabulary list, from 1,925,245 tokens to 822,875 or a decrease of 57.26\%. So that it can improve the quality of machine translation using the techniques we build. The translation quality increased up to 5 points BLEU from 40 to 45. The confidence level of the translation model or perplexity is also decreased from 2.5 to 0.62 (the smaller the number, the better).



\bibliography{anthology,acl2020,custom}
\bibliographystyle{acl_natbib}

\onecolumn
\appendix
\section{Appendices}
\label{sec:appendixA}
\begin{table*}[]

\resizebox{\textwidth}{!}{%
\begin{tabular}{@{}llll@{}}
\toprule
\textbf{No.} & \textbf{Type} & \textbf{Translation EN-ID Result Sentences} & \textbf{BLEU Score} \\ \midrule
1 & Our Result & \begin{tabular}[c]{@{}l@{}}
di dalam negeri , di$\sim$sumbang $\sim$kan 13 juta dolar untuk me$\sim$diri $\sim$kan \\ 
pe$\sim$teliti $\sim$an pe$\sim$teliti $\sim$an dan yayasan pe$\sim$teliti $\sim$an teluk di situ \\
saat itu me$\sim$sedia $\sim$kan ke$\sim$ber$\sim$ada $\sim$an yang di$\sim$sedia $\sim$kan \\
dari pe$\sim$duduk dan me$\sim$satu $\sim$kan fokus pada pe$\sim$teliti lain ter$\sim$masuk \\
pe$\sim$atur $\sim$an pe$\sim$iklan $\sim$an dan ke$\sim$dokter $\sim$an medis .
\end{tabular} & 29.69  \\
\cdashlinelr{2-4}
& Baseline Result & \begin{tabular}[c]{@{}l@{}}
Dalam hal ini , sekitar 13 juta dolar untuk mendirikan sebuah \\ sepenuhnya Fac dan surat kabar yang dimiliki Teluk Bay pada masa itu\\ pada saat itu teen oleh bintang dan Shaw Medali istana yang dilakukan\\
di lain yang Khan termasuk ler yang ada di historical akan Terletak di\\ historical akan .
\end{tabular} & 11.28  \\
 \cdashlinelr{2-4}
& Origin (EN) & \begin{tabular}[c]{@{}l@{}}
In 1987, Packard donated 13 million dollars to establish the Aquarium\\
of the Bay Monterrey research institute, and the Packard Foundation \\
at that time provided 90\% of McCombs and Shaw's expanded focus on\\
other researchers including setting an agenda on historical issues, \\ advertising and medical news.
\end{tabular} & -  \\
 \cdashlinelr{2-4}

 & Truth Ref (ID) & \begin{tabular}[c]{@{}l@{}}
Pada tahun 1987 , Packard menyumbangkan 13 juta dolar untuk mendirikan\\ lembaga penelitian Aquarium Teluk Monterrey , dan Packard Foundation \\ pada saat itu menyediakan 90 \% dari McCombs dan Shaw memperluas fokus\\ pada peneliti lain termasuk menetapkan agenda pada isu-isu sejarah , \\ iklan dan berita medis .
\end{tabular} & - \\
 \midrule
2 & Our Result & \begin{tabular}[c]{@{}l@{}}larva kumbang ini biasa $\sim$nya me$\sim$ bor ke dalam kayu dan \\
dapat me$\sim$ sebab $\sim$kan ke$\sim$ rusa $\sim$kan pada batang kayu hidup \\
 atau kayu yang telah di$\sim$ tebang
\end{tabular} & 69.44  \\
\cdashlinelr{2-4}
 & Baseline Result & \begin{tabular}[c]{@{}l@{}}Larva kumbang ini biasanya mengebor ke dalam kayu dan \\
 dapat menyebabkan kerusakan pada kayu dan kayu hidup \\
  yang telah alists .
\end{tabular} & 51.92  \\
 \cdashlinelr{2-4}
 
& Origin (EN) & \begin{tabular}[c]{@{}l@{}}
These beetle larvae usually drill wood and can cause\\
damage to live logs or felled wood.
\end{tabular} & -  \\
 \cdashlinelr{2-4}
 
 & Truth Ref (ID) & \begin{tabular}[c]{@{}l@{}}
 Larva kumbang ini biasanya mengebor kayu dan \\
 dapat menyebabkan kerusakan pada kayu gelondongan \\
 hidup atau kayu yang telah ditebang .
\end{tabular} & - \\
 \midrule

3 & Our Result & \begin{tabular}[c]{@{}l@{}}karena ber$\sim$ fungsi se$\sim$ bagai pe$\sim$ tutup salur $\sim$an yang lama \\
 dari sungai ini hanya me$\sim$ kandung air hitam karena seluruh \\
  air ber$\sim$ asal dari tangga rumah tangga dan industri .
\end{tabular} & 56.40 \\
\cdashlinelr{2-4}
 & Baseline Result & \begin{tabular}[c]{@{}l@{}}Karena 99 sebagai sebuah tribes dari Bahkan lama Sungai \\
  Cakung hanya membentuk air hitam karena seluruh air \\
   berasal dari tahun household dan Hong industri
\end{tabular} & 36.26  \\
 \cdashlinelr{2-4}
 
 & Origin (EN) & \begin{tabular}[c]{@{}l@{}}
Because it functions as a drainage, the old channel of\\
the Cakung River only contains black water because\\
all water comes from household and industrial waste.
\end{tabular} & -  \\
 \cdashlinelr{2-4}
 
 & Truth Ref (ID) & \begin{tabular}[c]{@{}l@{}}Karena berfungsi sebagai drainase , saluran lama Sungai \\
  Cakung hanya mengandung air hitam karena seluruh air \\
   berasal dari limbah rumah tangga dan industri .
\end{tabular} & - \\
 \midrule

\end{tabular}
}
\caption{Sample of translation result}
\label{translate}
\end{table*}
\label{sec:appendixB}
\begin{figure*}[htbp]
\centerline{\includegraphics[width=0.9\textwidth]{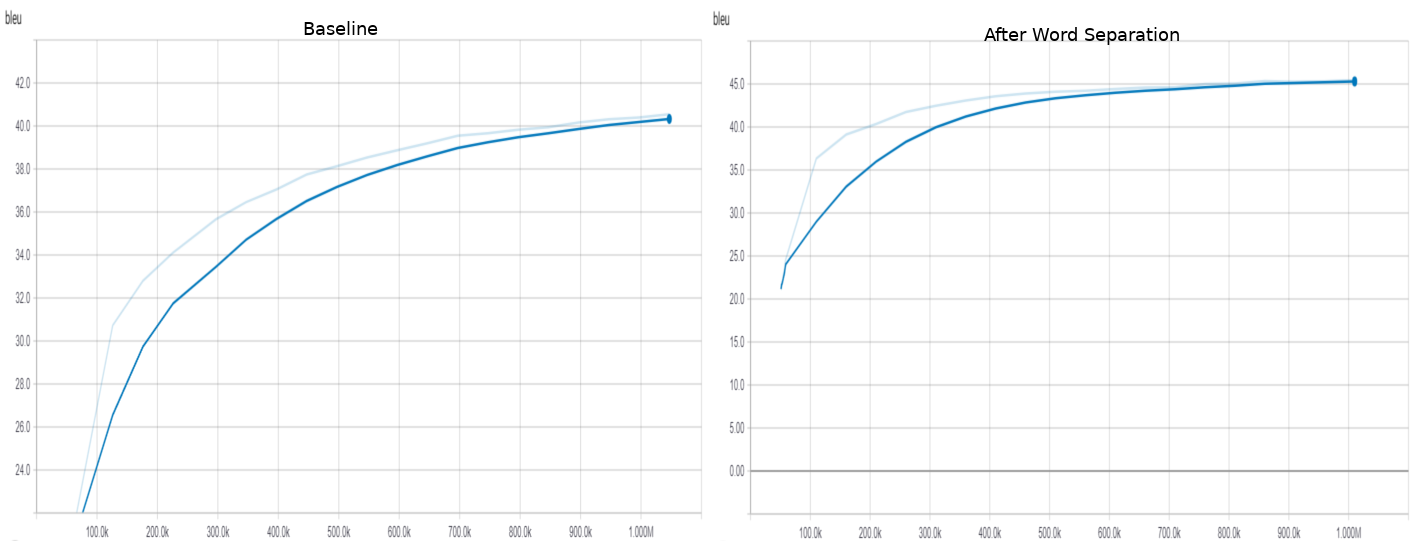}}
\caption{BLEU Score Improvement from 40 to 45}
\label{fig:4}
\end{figure*}
\begin{figure*}[htbp]
\centerline{\includegraphics[width=0.9\textwidth]{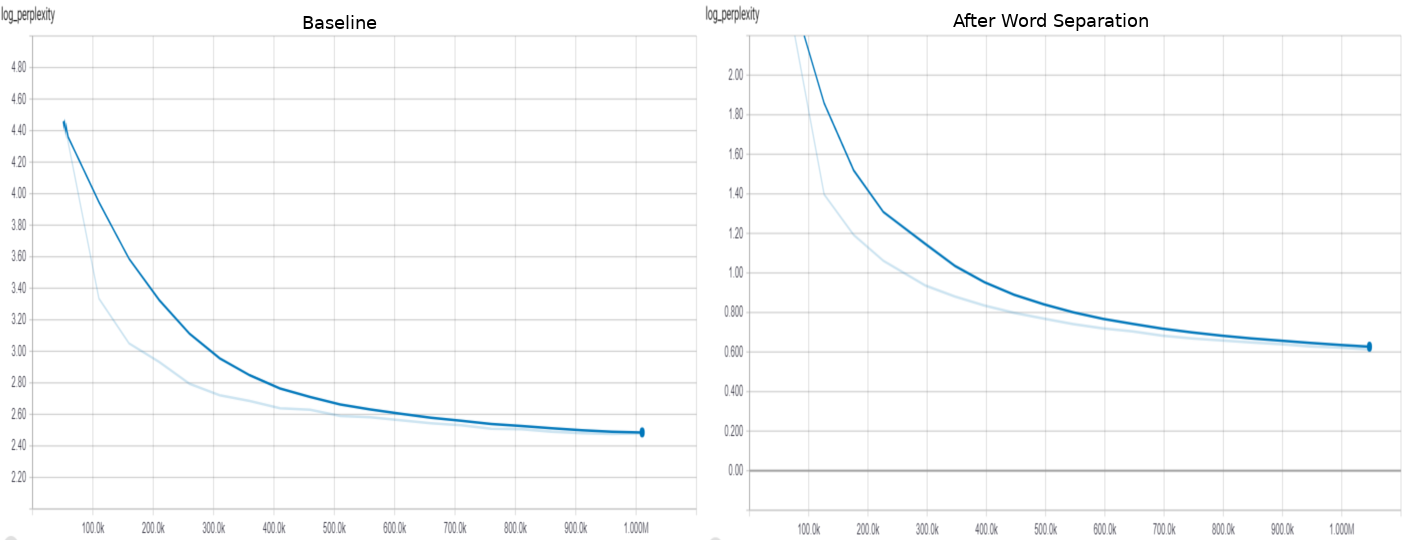}}
\caption{Log perplexity improvement from 2.5 to, 0.62}
\label{fig:5}
\end{figure*}

\end{document}